# Improved-Flow Warp Module for Remote Sensing Semantic Segmentation

*Yinjie Zhang, Yi Liu, Wei Guo*

*Abstract*—**Remote sensing semantic segmentation aims to assign automatically each pixel on aerial images with specific label. In this letter, we proposed a new module, called improved-flow warp module (IFWM), to adjust semantic feature maps across different scales for remote sensing semantic segmentation. The improved-flow warp module is applied along with the feature extraction process in the convolutional neural network. First, IFWM computes the offsets of pixels by a learnable way, which can alleviate the misalignment of the multi-scale features. Second, the offsets help with the low-resolution deep feature up-sampling process to improve the feature accordance, which boosts the accuracy of semantic segmentation. We validate our method on several remote sensing datasets, and the results prove the effectiveness of our method..**

## I. INTRODUCTION

WITH the development of remote sensors, remote sensing pictures have played an important role in urban planning, land-use resources statistics and e.t.c. High-resolution images change the traditional ways of surveying in geodesy and require in-time deciphering[1]–[3].

Recently, the rapid progress made by deep learning shines a light on semantic segmentation and achieves higher accuracy in many daily tasks.FCN [4] utilizes a full convolution network to assign each pixel a class and achieves better results than traditional algorithms. After FCN, UNET [5] applies encoder-decoder structure to maintain the high resolution output. SegNet [6] uses the mirror symmetric encoder-decoder to segment picture. To extract multi receptive field's feature and multi-resolution feature, DeepLab [7] series implement atrous convolution and ASPP module which boost the segmentation results. To cover up the global contextual information, GCN [8] enlarges the kernel size of convolution in neural network along with asymmetry structure to extract class-relation feature. As for the up-sample process, DeconvNet [9] modifies the decoder structure with the transposed convolution, which uses computable parameters recover the resolution of the feature map. Moreover, the feature pyramid network has been proposed to enhance the multi-resolution feature fusion, such as FPN [10], PPM [11], BiFPN [12], e.t.c.

However, remote sensing images are bird's eye view photoed and exist large scale variance. The complex land cover distribution accounts for the feature differences in single scale and the cross-scale spatial discrepancy, which hinders the improvement in segmentation results. EFCNet [13] modifies SENet [14] to fuse multi-scale features according to their channel level attention, which treats the scale fusion problem as a channel-wise selection. FarSeg [15] uses global contextual information, which concerns the foreground and background relationship, to enhance the related features and decrease the other objects. MUNET [16] adopts the UNET++ [17] structure to extract multi-resolution feature and fuse them through a simple concatenation and convolution process. To make the segmentation boundary more specific, EaNet [18] propose an edge-aware network for remote sensing images and creates LKPP module to acquire directional feature. D-LinkNet [19] incorporates transposed convolution with encoder-decoder structure to specific the detailed information of land-cover, which is the winner solution in DeepGlobe road extraction. MACANet [20] proposed AS-CEB module to learn different scale feature and recover the segmentation map with SAB block, which improves the results of very high resolution RS pictures.

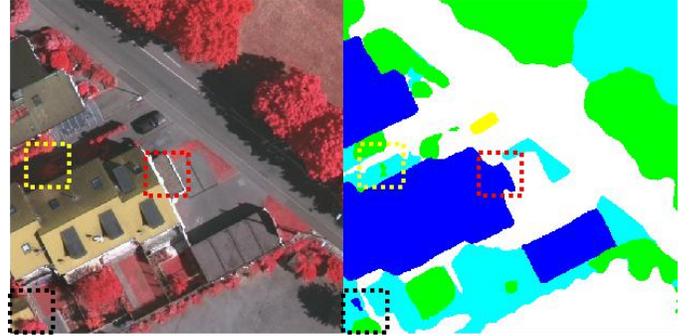

**Fig. 1.** Segmentation algorithm ignoring the differences in feature fusion causes the pixel prediction error.

Although previous works have made progress in multi-scale feature extraction and fusion, they ignore a basic yet important question, which cross-scale feature may not be spatial registered, which means after the convolution down sampling the feature in deeper network may not be accordance with other layers' feature. This kind of discrepancy may cause the intraclass differences and increase the interclass similarity, which hinders the pixel-wise class assignment.

In this letter, we propose an effective way, improved-flow warp module(IFWM), to eliminate the spatial discrepancy in remote sensing images segmentation. The main contribution are summarized as follows: First, the high-resolution network learns the features from different scales. The multi branch structure helps the network maintain the feature from different resolution. Second, the IFWM plays the coordinate-correction role in the feature map fusion process, improving the spatial accordance of feature map during the feature fusion process. Extensive experiments demonstrates the effectiveness of our method.

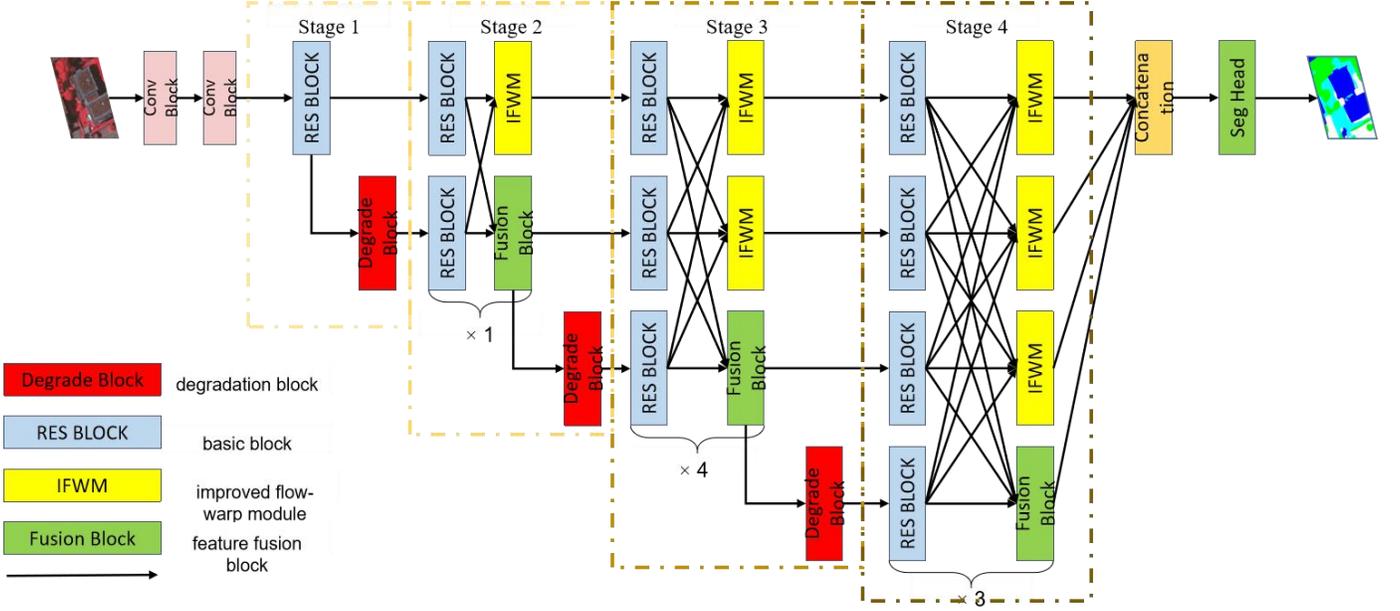

**Fig. 2.** The Structure of IFWM Network. The proposed IFWM module helps the small-spatial-size deep feature align with the shallow feature.

## II. METHODOLOGY

Fig. 2 illustrates the overall structure of the proposed network, the high-resolution network is taken as the backbone to extract multi-scale feature.Meanwhile, the IFWM serves as a " adjust-function " in cross-scale feature fusion process, which helps the low-resolution feature maps register with the high-resolution map. Then a final layer fuse the feature and output the prediction.

### A. High-Resolution Network

High-resolution network maintains different scale feature maps and exchanges the contextual information across scales.The network can be divided into 3 blocks, the feature extraction block, the feature fusion block and the resolution degradation block. The input image goes through some convolution blocks to reduce the feature resolution and goes into the feature extraction block. The network adopts residual block to extract feature. In order to obtain different resolution feature, the feature map is sent to resolution degradation block, which consists of $3 \times 3$ convolution blocks. To fuse the different scale contextual information, the feature fusion block uses $1 \times 1$ convolution blocks and bi-linear up-sample adapt deep feature to the shallow and uses $3 \times 3$ convolution blocks with stride 2 to adapt shallow feature to the deep. After the 4 different resolution feature extraction, the network concatenate the map and make the prediction.

### B.Improved Flow Warp Module

The improves flow warp module can learn the warp-map of low-resolution feature and make the cross-scale fusion process more reasonable. The IFWM works in the feature fusion process. It is inspired by SFNet [21]. The structure of IFWM is depicted in Fig. 3.

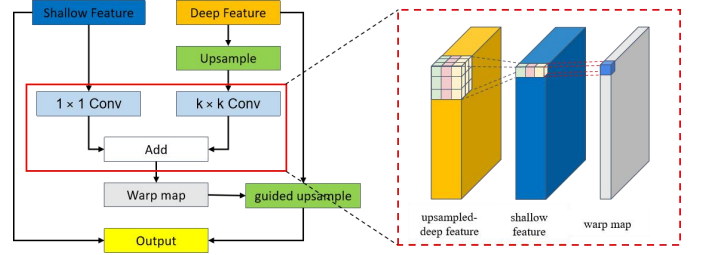

**Fig. 3.** The Implementation of IFWM Module

Different from SFNet, to achieve the warp-map that guides the up-sample process, IFWM uses different receptive fields for different resolution feature map, rather than an unified 3*3 convolution block. In our theory, the feature fusion process in up-sampling is to make accordance between one pixel in shallow feature and 4 pixels in deep feature. The core task for spatial register is to select proper contextual information representation from deep feature for the shallow feature. Under this consumption, we directly use 1*1 convolution block for the shallow feature. While for the deeper feature, we assign different receptive fields by the k*k convolution block. The kernel size is accordance with the feature map ratio between the deep and shallow, which is set to make the spatial register between one pixel in shallow and 4 pixels in deep. To fulfill this task, the kernel sizes of different scales are 3, 7 and 15, in which 3 is for 2 times up-sampling and 15 is for 8 times up-sampling. After the convolution block, the warp-map is computed by directly adding and then guides the point-flow for the deep feature, which means up-sampling under the direction of calculated weights of pixels.

The computation process can be summarized as following equations.

$$\text{warp\_map} = \text{pixel\_conv}(X_S) + \text{region\_conv}(X_D) \quad (1)$$

Where the pixel_conv means 1*1 convolution for shallow feature and region_conv means k*k convolution for deep feature. $X_S$ represents shallow feature and $X_D$ represents deep feature.

After the warp-map has been calculated, the fusion process uses the grid sample function and adds at last.

$$output = X_S + grid\_sample(warp\_map, X_D) \quad (2)$$

The grid sample function is a flexible sample function which changes the uniform sample process and replaces it with calculable weights. The computation process can be depicted as follows.

$$F(P_m) = \sum_{p \in N(P_R)} W_P F(P_o) \quad (3)$$

$N(P_R)$ means the four corner pixels near the target pixels, $W_P$ means the the bi-linear weights estimated by the distance of

TABLE I
PERFORMANCE ON THE ISPRS VAIHINGES DATASET

| Method | Building | Low Veg. | Tree | Car | Imp.Surf. | mF1 | PA | mIOU |
|--------|----------|----------|------|-----|-----------|-----|-----|------|
| Unet | 76.94 | 84.56 | 60.56 | 73.29 | 61.25 | 82.92 | 84.80 | 71.32 |
| PSPNet | 80.08 | 87.37 | 62.41 | 74.76 | 71.88 | 85.65 | 86.40 | 75.30 |
| FCN | 80.09 | 87.65 | 62.43 | 74.80 | 67.96 | 85.15 | 86.44 | 74.59 |
| DeepLab V3+ | 80.19 | 88.15 | 62.46 | 74.84 | 69.66 | 85.47 | 86.50 | 75.06 |
| HRnet | **80.72** | **89.17** | 62.58 | 75.32 | 74.14 | 86.33 | 86.92 | 76.39 |
| IFWM | 80.50 | 88.72 | **63.63** | **75.42** | **74.99** | **86.54** | **86.95** | **76.65** |

TABLE II
PERFORMANCE ON THE WHU BUILDING DATASET

| Method | Backbone | Precision | Recall | F1 | IoU |
|--------|----------|-----------|--------|-----|-----|
| Unet | resnet50 | 0.93 | 0.94 | 0.93 | 87.53 |
| PSPNet | resnet50 | 0.93 | 0.94 | 0.93 | 88.60 |
| FCN | resnet50 | 0.93 | 0.94 | 0.93 | 87.42 |
| DeepLab V3+ | resnet50 | 0.94 | 0.95 | 0.94 | 89.42 |
| HRnet | hrnet | 0.94 | 0.95 | 0.94 | 89.68 |
| IFWM | hrnet | 0.94 | 0.95 | 0.94 | **90.15** |

warped grid. Different from the uniform sampling process, the weights are not directly the scale ratio but calculated by the feature from both deep feature and shallow feature.

## III. EXPERIMENTS

### A. Datasets

Our experiments conduct on the ISPRS Vaihingen semantic labeling benchmark and the WHU Building dataset. The ISPRS Vaihingen dataset consists of 33 aerial orthoimage with ground sample distance of 9 cm. Each image has near-infrared (NIR), red (R), and green (G) band and corresponding DSM data. There are 16 images with labels , a total of six classes. We use 11 images for training and 5 images (id: 11, 15, 28, 30, and 34) for testing. We use sliding windows with 400 pixels stride to extract $512 \times 512$ patches and then augment the patches with randomly rotation. WHU Building dataset consists of 8189 $512 \times 512$ tiles with 0.3 m GSD, of which 4736 for training, 1036 for validation and 2416 for testing. WHU Building has 3 bands and two labels, the building and the background.

### B. Implementation Details

The proposed model is implemented with PyTorch. The training platform is Nvidia-Tesla-V100 16GB. We adopt the ImageNet pretrained parameters to initialize the architecture and fine-tune the module with IFWM on our 2 datasets. We use SGD optimizer with a initial learning rate of 0.01. Meanwhile, we use an exponentially decaying learning rate with a decay rate of 0.9 in all epochs. The training epoch is set to 100 for ISPRS Vaihingen dataset and 80 for WHU Building dataset. The batch size is set to 8 to all the experiments in this letter. For ISPRS Vaihingen dataset, we only use the top-image and do not introduce the DSM in training and predicting process.

### C. Evaluation Metrics

We use 4 quantitative indexes to evaluate the effectiveness of our method: F1-score, Pixel Accuracy, Intersection over Union and mean Intersection over Union (mIoU). The formulations are defined as follows:

$$Precision = \frac{TP}{TP+FP} \quad Recall = \frac{TP}{TP+FN} \quad (4)$$

$$F_1 = 2 \cdot \frac{Precision \cdot Recall}{Precision + Recall} \quad (5)$$

$$IoU = \frac{TP}{TP+FP+FN}, \quad mIoU = \frac{1}{n}\sum_{i=0}^{n} IoU_i \quad (6)$$

$$PA = \frac{\sum_{i=0}^{n} p_{ii}}{\sum_{i=0}^{n}\sum_{j=0}^{n} p_{ij}} \quad (7)$$

TP means true positive prediction, TN means true negative prediction, FP means false positive prediction and FN means false negative prediction.

### D. Experimental Results

In order to evaluate the IFWM framework, we compare the performance of other models, UNET [5], PSPNET [11], FCN-8s [4], DeepLab V3+ [7]. As displayed in Table I and Table II, the IFWM achieves the highest IOU in WHU Building dataset and highest mF1, PA and mIoU in ISPRS Vaihingen dataset.

In ISPRS Vaihingen dataset, compared to HRNET, the mF1, PA and mIoU are increased by 0.21%, 0.03%, 0.26% respectively. In all classes, the proposed IFWM improves Trees and Imp.Surf. Significantly, which achieves 1.05% and 0.85% improvement. The IFWM decreases in building and low vegetation slightly, this mainly stems from the boundary between building and vegetation, where the cover-up region is hard to distinguish in remote sensing image. The details is visualized in *Visualization* part. The improvement in PA means the IFWM leads more pixels to their actual labels successfully, by the means of adjusting the feature assignment specifically.

In WHU Building dataset, due to the limitation of reserved digit, the performance can not be captured through precision, recall or F1-score. However, the IOU shows the effectiveness of IFWM, which improves 0.47% compared to HRNET. Different from ISPRS Vaihingen dataset, we use IOU to evaluate the performance in building extraction rather than mIOU which considers the accuracy in background. The highest IOU score shows that the IFWM can extract the building more accurately due to the appropriate feature adjustment for the structural shape in building class.

Alleviate the holes in large region and improve the boundary of different classes. Strengthen the small region and eliminate some wrong prediction in buildin

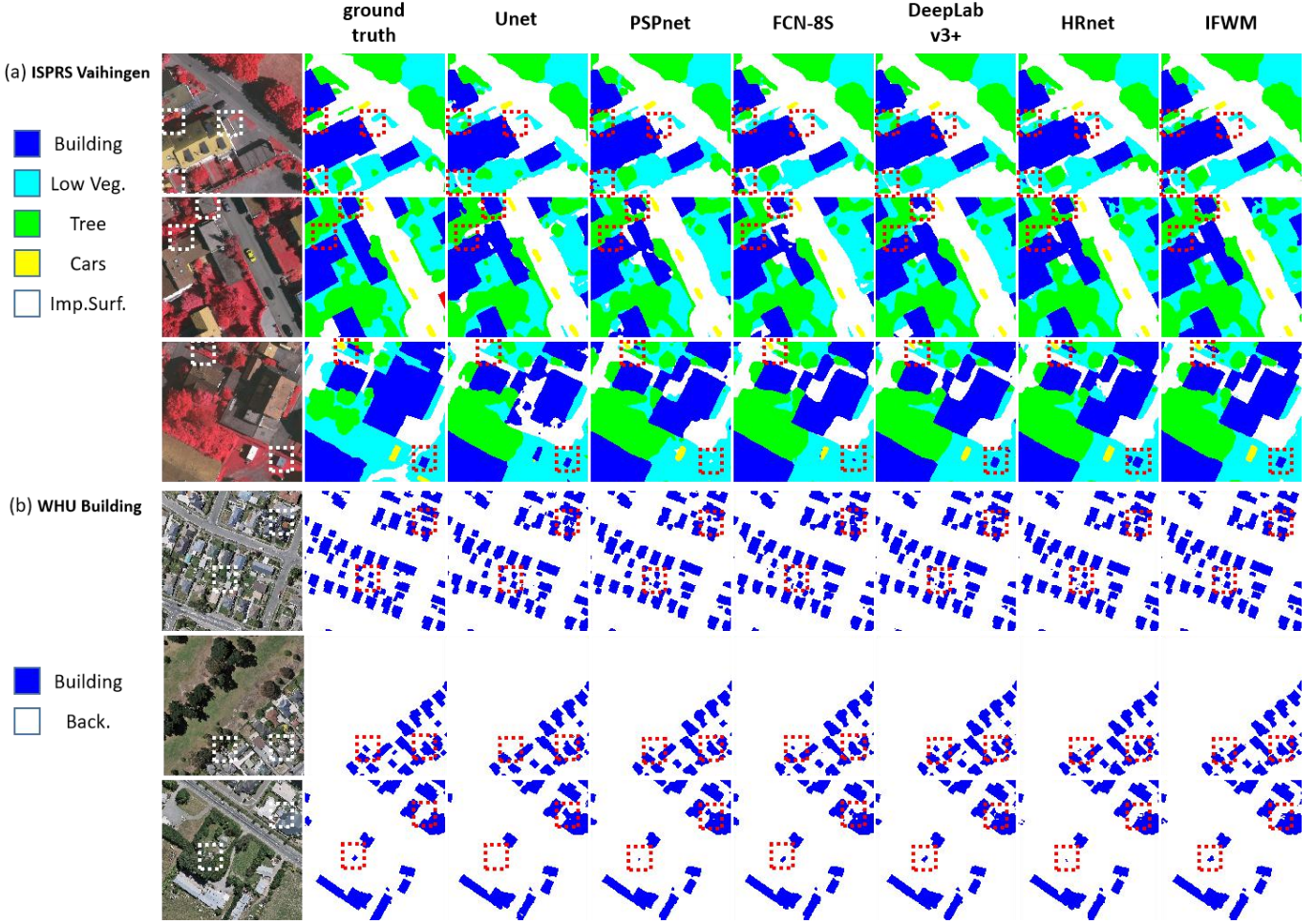

**Fig. 4.** Visualization of experiment conducted on 2 datasets

*E.Visualization*

To better show the performance of IFWM and the improvement in details, we select 3 samples in 2 datasets for visualization, as shown in Fig. Rows 1 to Rows 3 in Fig. 4 a are the visualization results of ISPRS Vaihingen dataset. The first row proves that the IFWM can predict building region more precisely, especially at the boundary, which shows straight lines rather than curves. This is because the IFWM introduces the feature alignment to adjust the feat. In the second row, due to the similarities in Low Veg. and Trees, the other methods can not distinguish them in predictions, while the IFWM can avoid this by selectively choosing the feature from different scales. The third row shows that IFWM can help the structure not to ignore small region or object in the image. This is mainly because the adjust in feature across different scales to avoid feature encroaching.

In Fig. 4 b are the visualization results in WHU Building dataset. The first and second row show the better recall performance in building extraction. There exist some small building on the edge of architectural complex. In other ways, the buildings on the edge can be easily ignored and thus lead to prediction error. The IFWM can predict this kind of buildings correctly. The third row shows that the quality of prediction in boundaries of building. Owing to the feature adjustment across different scales by IFWM, the edge can be saved almost to the ground truth and the input image, which makes the prediction right.

*F.Ablation Study*

To verify the effectiveness of the large kernel in IFWM and the effectiveness of computing the offset separately, we compare the ablation analysis on the ISPRS Vaihingen dataset. The image is segment to 256 × 256 size to fasten the computation.

Listed in Table III, the calculation process is the function how the warp map achieved. Concat+conv menas concatenating the deep and shallow feature and using unifoem convolution to compute the weights. Conv+add is computing warp map separately for deep and shallow feature, and adding the results. The kernel size column means convolution size for shallow feature and deep feature.

TABLE III
ABLATION STUDY

| Method | Calculation process | Kernel size | mF1 | PA | mIOU |
|--------|---------------------|-------------|------|------|------|
| HRnet | × | × | 85.42 | 86.03 | 74.97 |
| SF | concat + conv | 3×3 | 85.26 | 85.93 | 74.73 |
| LSF | concat + conv | k×k | 85.65 | 86.34 | 75.31 |
| RIFW | conv + add | 3×3+k×k | 85.50 | 86.24 | 75.09 |
| IFWM | conv + add | 1×1+k×k | **85.76** | **86.38** | **75.48** |

The experiment results are shown in Table III. SF means using SF module directly in the same location as IFWM. LSF means enlarge the receptive field for the concatenated feature as IFWM. RIFW means using 3×3 convolution for the shallow feature and k×k for the deep feature. IFWM is the proposed method. As shown in the table, directly using SF introduces uncertainties which makes it hard to converge, so the results appear inferior to the basic architecture. Using large kernel to replace the 3×3 convolution blocks makes improvement but introduces large computation cost. The RIFW shows that considering more pixels in shallow feature is not necessary and may decrease the performance of the k×k convolution. The IFWM shows the best results among all the other results. These experimental results verify the effectiveness of the IFWM module.

## IV. CONCLUSION

In this letter, we proposed an end-to-end remote sensing semantic segmentation network, which utilizes HRNet and the IFWM module to extract different scale feature and adjust the feature in spatial dimension, thus reduce the misalignment in feature to improve the segmentation results. Through multiple experiments, the proposed method outperform other models in different datasets. Through ablation study, we verify the effectiveness of IFWM in feature fusion process. However, the IFWM only considers the alignment while still fuses the feature by adding. How to select and align the feature simultaneously is our future work.

## ACKNOWLEDGMENT

The numerical calculations in this paper have been done on the supercomputing system in the Supercomputing Center of Wuhan University.